\documentclass{article}

 \usepackage[final,nonatbib]{neurips_2019}

\usepackage[utf8]{inputenc} 
\usepackage[T1]{fontenc}    
\usepackage{hyperref}       
\usepackage{url}            
\usepackage{booktabs}       
\usepackage{amsfonts}       
\usepackage{nicefrac}       
\usepackage{microtype}      
\usepackage{bm}
\usepackage{amsmath}
\usepackage{amssymb}
\usepackage[lined,boxed,commentsnumbered, ruled,linesnumbered]{algorithm2e}
\usepackage{algorithmic}
\usepackage{multirow}
\usepackage{stmaryrd}
\usepackage{graphicx}
\usepackage{epstopdf}
\newcommand{\bs}[1]{{\bm{#1}}}
\newcommand{\encrypt}[1]{{\llbracket #1 \rrbracket}}

\title{A Quasi-Newton Method Based Vertical Federated Learning Framework for Logistic Regression}

\author{%
  Kai Yang \\
  WeBank \& ShanghaiTech University \\
  \texttt{yangkai@shanghaitech.edu.cn}\\
  \And
  Tao Fan \\
  WeBank \\
  \texttt{dylanfan@webank.com}\\
  \And
  Tianjian Chen \\
  WeBank \\
  \texttt{tobychen@webank.com}\\
  \And
  Yuanming Shi \\
  ShanghaiTech University \\
  \texttt{shiym@shanghaitech.edu.cn} \\
  \And 
  Qiang Yang \\
  Hong Kong University of Science and Technology \\
  \texttt{yangqiang@hkust.edu.cn} \\
}

\begin{document}

\maketitle

\begin{abstract}
Data privacy and security becomes a major concern in building machine learning models from different data providers. Federated learning shows promise by leaving data at providers locally and exchanging encrypted information. This paper studies the vertical federated learning structure for logistic regression where the data sets at two parties have the same sample IDs but own disjoint subsets of features. Existing frameworks adopt the first-order stochastic gradient descent algorithm, which requires large number of communication rounds. To address the communication challenge, we propose a quasi-Newton method based vertical federated learning framework for logistic regression under the additively homomorphic encryption scheme. Our approach can considerably reduce the number of communication rounds with a little additional communication cost per round. Numerical results demonstrate the advantages of our approach over the first-order method.
\end{abstract}

\section{Introduction}
With the surge of artificial intelligence (AI) driven services including recommender system and natural language processing, data privacy and security have raised worldwide concerns \cite{yang2019federated}. More and more stringent requirements of data privacy and security become an emerging trend of laws and regulations from states across the world. A known example is the General Data Protection Regulation (GDPR) by the European Union \cite{albrecht2016gdpr}. Traditional AI service providers usually collect and transfer data instances from one party to anther party. Then a machine learning model is trained at the cloud data center with the fused data set. However, it faces challenges of data breach and violation of data protection laws and regulations \cite{wiki:facebookdatabreach}. 

Recently, federated learning \cite{yang2019federated,konevcny2016federated,mcmahan2017communication} is an emerging frontier field studying privacy-preserving collaborative machine learning while leaving data instances at their providers locally. A line of works \cite{konevcny2016federated,mcmahan2017communication,yang2018federated} focus on the horizontal structure, in which each node has a subset of data instances with complete data attributes. There are also many researches studying the vertical federated learning structure where the data set is vertically partitioned and owned by different data providers. That is, each data provider holds a disjoint subset of attributes for all data instances. The target is to learn a machine learning model collaboratively without transferring any data from one data provider to another. In particular, \cite{cheng2019secureboost} proposes a privacy-preserving tree-boosting system \textit{SecureBoost} and \cite{hardy2017private} propose a logistic regression framework for vertically partitioned data.

Communication is one of the main bottlenecks in federated learning due to the much worse network conditions than the cloud computing center \cite{konevcny2016federated}. To address the communication challenge in horizontal federated learning, structured updates are considered in \cite{konevcny2016federated} to reduce the communication costs per round and an iterative model averaging algorithm is proposed in \cite{mcmahan2017communication} to reduce the number of communication rounds. For vertical federated learning structure, \cite{hardy2017private} considers a two party (denoted by party A and party B) logistic regression problem and proposes a stochastic gradient descent (SGD) method based privacy-preserving framework. Due to the slow convergence of first-order algorithms, it requires a large number of communication rounds. This work shall propose a quasi-Newton method based vertical federated learning system with sub-sampled Hessian information to reduce the communication round.

\paragraph{Related Works}
Second-order Newton's method is known to converge faster than first-order gradient based methods. To avoid the high cost of computing the inversion of Hessian matrix, a well recognized quasi-Newton method Limited-memory BFGS (L-BFGS) \cite{nocedal2006numerical} algorithm is proposed by directly approximating inverse Hessian matrix. There are a number of works \cite{schraudolph2007stochastic,byrd2016stochastic,moritz2016linearly} focus on developing stochastic quasi-Newton algorithms for problems with large amounts of data. However, the inverse Hessian estimated by \cite{schraudolph2007stochastic} may be not stable for small batch sizes and the algorithm in \cite{moritz2016linearly} requires computing the full gradient which would double the communication cost in each epoch compared with SGD. This paper develops a communication efficient vertical federated learning framework based on the stochastic quasi-Newton method proposed in \cite{byrd2016stochastic}.


\section{Problem Statement}
Consider a typical logistic regression problem with vertically partitioned data \cite{hardy2017private}. Let $\bs{X}\in\mathbb{R}^{n \times T}$ be the data set consisting of $T$ data samples and each instance has $n$ features. The class attribute information, i.e., the label of data, is given by $\bs{y}\in\{-1,+1\}^{T}$. The data set is vertically partitioned and distributed on two honest-but-curious private parties A (the \textbf{host} data provider with only features) and B (the \textbf{guest} data provider with features and labels). Let $\bs{X}^A\in\mathbb{R}^{n_A \times T}$ be the data set owned by party A and $\bs{X}^B\in\mathbb{R}^{n_B \times T}$ owned by party B. Each party owns a disjoint subset of data features over a common sample IDs with $\bs{X}=(\bs{X}^A,\bs{X}^B)$. In addition, only party B has access to the labels $\bs{y}$. 
The target of logistic regression is to train a linear model for classification by solving
\vspace{-0.5em}\begin{equation}
\mathop{\textrm{minimize}}_{\bs{w}\in\mathbb{R}^{n}}\quad \frac{1}{T}\sum_{i}^{T}l(\bs{w};\bs{x}_i,y_i),
\end{equation}
where $\bs{w}$ is the model parameters, $\bs{x}_i$ is the $i$-th data instance and $y_i$ is the corresponding label. The negative log-likelihood loss function is given by
	$l(\bs{w};\bs{x}_i,y_i)=\log(1+\exp(y_i\bs{w}^{\sf{T}}\bs{x}_i))$.
In this paper, we suppose that party $A$ and party $B$ hold the model parameters corresponding to their features respectively, which can be denoted as $\bs{w}=(\bs{w}^A,\bs{w}^B)$ where $\bs{w}^A\in\mathbb{R}^{n_A}$ and $\bs{w}^B\in\mathbb{R}^{n_B}$. 


\cite{hardy2017private} proposes a stochastic gradient descent (SGD) based vertical logistic regression framework by computing gradients via exchanging encrypted intermediate values at each iteration. Specifically, party $A$ and party $B$ collaboratively compute the vertically partitioned encrypted gradient $\bs{g}^A\in\mathbb{R}^{n_A}$ and $\bs{g}^B\in\mathbb{R}^{n_B}$, which can be decrypted by the third party. To achieve secure computation without transferring data from one party to another, the additively homomorphic encryption is adopted. Additively homomorphic encryption schemes such as Paillier \cite{paillier1999public} allow any party can encrypt their data with a public key, while the private key for decryption is owned by the third party, i.e., the \textbf{coordinator}. With additively homomorphic encryption we can compute the additive of two encrypted numbers as well as the product of an unencrypted number and an encrypted one, which can be denoted as
$  \encrypt{u}+\encrypt{v} = \encrypt{u+v},  v\cdot \encrypt{u} = \encrypt{vu}$
by using $\encrypt{\cdot}$ as the encryption operation.
Unfortunately, the loss function and its gradient cannot be computed directly with additively homomorphic encryption. To address this issue, we will adopt the Taylor approximation for the loss function is proposed in \cite{hardy2017private,aono2016scalable} as
\begin{equation}\label{eq:taylor_loss}
\textrm{Taylor loss:}\quad	l(\bs{w};\bs{x}_i,y_i) \approx \log2-\frac{1}{2}y_i\bs{w}^{\sf{T}}\bs{x}_i+\frac{1}{8}(\bs{w}^{\sf{T}}\bs{x}_i)^2.
\end{equation}

\section{A Quasi-Newton Method Based Vertical Federated Learning Framework}
\begin{figure}[h]
    \centering
    \includegraphics[width=0.7\columnwidth]{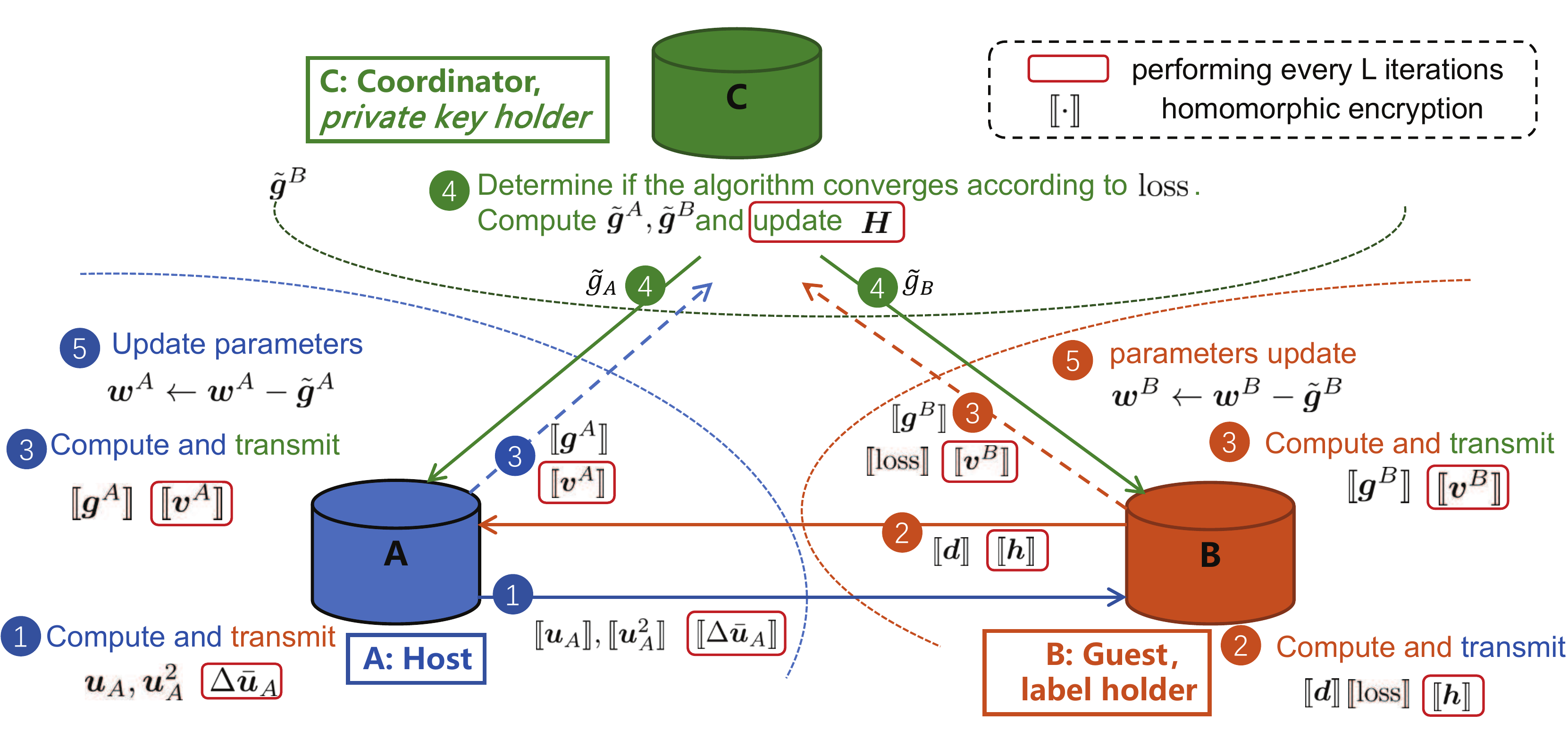}
    \caption{A Quasi-Newton Framework for Vertical Federated Learning}
    \label{fig:sqn}
\end{figure}

In federated learning, the communication cost between different parties is much more expensive than it in the cloud computing center since the data providers are usually across distant data centers, across different networks, or even in a wireless environment with limited bandwidth \cite{mcmahan2017communication}. So it becomes one of the main bottlenecks for efficiently model training. For this reason, we develop a communication efficient vertical federated learning framework by  incorporating second-order information \cite{byrd2016stochastic} to reduce the communication rounds between parties, which is illustrated in Fig. \ref{fig:sqn}.


The gradient and the Hessian of the Taylor loss in equation (\ref{eq:taylor_loss}) with respect to the $i$-th data instance are respectively given by
	$\nabla l(\bs{w};\bs{x}_i,y_i) \approx \left(\frac{1}{4}\bs{w}^{\sf{T}}\bs{x}_i-\frac{1}{2}y_i\right)\bs{x}_i, \nabla^2 l(\bs{w};\bs{x}_i,y_i) \approx \frac{1}{4}\bs{x}_i\bs{x}_i^{\sf{T}}$.
In the $k$-th iteration, classical L-BFGS algorithm uses the history information in last $M$ iterations by differencing gradient and model parameters between every two consecutive iterations to obtain an estimated inverse Hessian matrix $\bs{H}\in\mathbb{R}^{n\times n}$. But it will lead to a unstable curvature estimation if we use mini-batch data instead of full data. To obtain a stable estimation of $\bs{H}$, we shall use the sub-sampled Hessian information as suggested by \cite{byrd2016stochastic}. Moreover, the curvature information $\bs{H}$ can be updated every $L$ iterations to reduce the communication overhead as well as improve the stability of quasi-Newton algorithm. The details of computing the key ingredients for our system are introduced in the following part.

\paragraph{Computing Loss and Gradient at Party A\&B}
Let $\mathcal{S}\subseteq\{1,\cdots,T\}$ be the index set of the chosen mini-batch data instances. The corresponding loss and gradient are given by
	$\textrm{loss}=F(\bs{w})=\frac{1}{|\mathcal{S}|}\sum_{i\in\mathcal{S}}l(\bs{w};\bs{x}_i,y_i),\quad\bs{g}=\nabla F(\bs{w})=\frac{1}{|\mathcal{S}|}\sum_{i\in\mathcal{S}}\nabla l(\bs{w};\bs{x}_i,y_i) $.
By denoting $\bs{u}_A=\{\bs{u}_A[i]={\bs{w}^A}^{\sf{T}}\bs{x}_i^A:i\in\mathcal{S}\},\bs{u}_A^2=\{\bs{u}_A^2[i]=({\bs{w}^A}^{\sf{T}}\bs{x}_i^A)^2:i\in\mathcal{S}\}$ for party A (similarly $\bs{u}_B$ and $\bs{u}_B^2$ for party B) and $\bs{d}=\{d_i:i\in\mathcal{S}\}$, the encrypted loss and gradient can be computed by transmitting $\encrypt{\bs{u}_A}$ from party A to party B, and transmitting $\encrypt{\bs{d}}$ from B to A following
\begin{align}
    &\encrypt{\textrm{loss}}\approx \frac{1}{|\mathcal{S}|}\!\sum_{i\in\mathcal{S}}\encrypt{\log2}-\frac{1}{2}y_i(\encrypt{\bs{u}_A[i]}+\encrypt{\bs{u}_B[i]})+\frac{1}{8}(\encrypt{\bs{u}_A^2[i]}+2\bs{u}_B[i]\encrypt{\bs{u}_A[i]}+\encrypt{\bs{u}_B^2[i]}) \label{eq:loss}\\
    &\encrypt{\bs{g}}\approx \frac{1}{|\mathcal{S}|}\sum_{i\in\mathcal{S}}\encrypt{d_i}\bs{x}_i=(\underbrace{\sum_{i\in\mathcal{S}}\encrypt{d_i}\bs{x}_i^A}_{\encrypt{\bs{g}^A}},\underbrace{\sum_{i\in\mathcal{S}}\encrypt{d_i}\bs{x}_i^B}_{\encrypt{\bs{g}^B}}),~\encrypt{d_i}=\frac{1}{4}(\encrypt{\bs{u}_A[i]}+\encrypt{\bs{u}_B[i]}+\encrypt{-\frac{1}{2}y_i}).\label{eq:gradient}
\end{align}

\paragraph{Computing Updates for Estimating Curvature Information at Party A\&B} To achieve cheap communication costs introduced additionally, the curvature information $\bs{H}$ is updated every $L$ iterations at the coordinator by collecting encrypted $\bs{v}=(\bs{v}^A,\bs{v}^B)\in\mathbb{R}^n$ from party A and B. Specifically, every $L$ iterations we shall compute the difference of average model parameters as
\begin{equation}\label{eq:s}
	\bs{s}_t = \bar{\bs{w}}_{t}-\bar{\bs{w}}_{t-1}=(\bs{s}_t^{A}, \bs{s}_t^{B}),~\bar{\bs{w}}_t=\!\!\sum_{i=k-L+1}^{k}\!\!\bs{w}_i/L,~\bar{\bs{w}}_{t-1}=\!\!\sum_{i=k-2L+1}^{k-L}\!\!\bs{w}_i/L
\end{equation}
at party $A$ and party $B$. Then the product of sub-sampled Hessian $\nabla^2 \hat{F}(\bar{\bs{w}}_t)$ and average model difference $\bs{s}_t$ are given by
\begin{equation}\label{eq:v}
 	\bs{v}_t=\nabla^2 \hat{F}(\bar{\bs{w}}_t)\bs{s}_t,~\textrm{where}~\nabla^2 \hat{F}(\bar{\bs{w}}_t)=\frac{1}{|\mathcal{S}_H|}\sum_{i\in\mathcal{S}_H}\nabla^2 l(\bar{\bs{w}}_t;\bs{x}_i,y_i)=\frac{1}{|\mathcal{S}_H|}\sum_{i\in\mathcal{S}_H} \bs{x}_i\bs{x}_i^{\sf{T}}.
 \end{equation} 
The sub-sampled Hessian is calculated with respect to a randomly chosen subset of data $\mathcal{S}_H$. Under additively homomorphic encryption, $\encrypt{\bs{v}_t}$ can be computed following
\begin{equation}
    \encrypt{\bs{v}_t}=\frac{1}{|\mathcal{S}_H|}\sum_{i\in\mathcal{S}_H}\encrypt{h_i}\bs{x}_i=(\encrypt{\bs{v}_t^A},\encrypt{\bs{v}_t^B})=\Big(\frac{1}{|\mathcal{S}_H|}\sum_{i\in\mathcal{S}_H}\encrypt{h_i}\bs{x}_i^A,\frac{1}{|\mathcal{S}_H|}\sum_{i\in\mathcal{S}_H}\encrypt{h_i}\bs{x}_i^B\Big),
\end{equation}
where $h_i=\Delta\bar{u}_i^A+\Delta\bar{u}_i^B={\bs{s}_t^A}^{\sf{T}}\bs{x}_i^{A}+{\bs{s}_t^B}^{\sf{T}}\bs{x}_i^{B}$. By transmitting $\encrypt{\Delta\bar{\bs{u}}_A}=\{\encrypt{\Delta\bar{u}_i^A}:i\in\mathcal{S}_H\}$ from party A to party B, and transmitting $\encrypt{\bs{h}}=\{\encrypt{h_i}:i\in\mathcal{S}_H\}$ from B to A,  the corresponding components $\encrypt{\bs{v}_t^A}$ can be computed at party A and $\encrypt{\bs{v}_t^B}$ is computed at party B privately. 

\paragraph{Computing Descent Direction at the Coordinator}
After collecting the encrypted loss $\encrypt{\textrm{loss}}$, gradient $\encrypt{\bs{g}}$, and $\encrypt{\bs{v}}$ from party A\&B, the coordinator should determine a descent direction $\tilde{\bs{g}}$ for updating $\bs{w}^A$ and $\bs{w}^B$, i.e., $\bs{w}\leftarrow \bs{w}-\tilde{\bs{g}}=(\bs{w}^A -\tilde{\bs{g}}^A,\bs{w}^B-\tilde{\bs{g}}^B)$. Given an estimated $\bs{H}$, the descent direction is given by $\tilde{\bs{g}}=\eta \bs{H}\bs{g}$ where $\eta>0$ is the learning rate. Every $L$ iterations, $\bs{v}$ and $\bs{s}$ are stored in two queues with length $M$. $\bs{H}$ is determined by successively computing 
\begin{equation}\label{eq:H}
	\bs{H}\leftarrow(\bs{I}-\rho_j\bs{s}_j\bs{v}_j^{\sf{T}})\bs{H}(\bs{I}-\rho_j\bs{v}_j\bs{s}_j^{\sf{T}})+\rho_j\bs{s}_j\bs{s}_j^{\sf{T}},~\rho_j=1/(\bs{v}_j^{\sf{T}}\bs{s}_j), \forall j=t-M+1,\cdots,t
\end{equation}
from the initial point $\bs{H}=(\bs{v}_t^{\sf{T}}\bs{s}_t/\bs{v}_t^{\sf{T}}\bs{v}_t)\bs{I}$.
It should be noted that $\bs{s}_t$ can be computed locally at the coordinator as $\bs{s}_t=\sum_{i=k-L+1}^{k}\tilde{\bs{g}}_i/L-\sum_{i=k-2L+1}^{k-L}\tilde{\bs{g}}_i/L$ without any additional transmissions. The overall quasi-Newton method based vertical federated learning framework is illustrated in Fig. \ref{fig:sqn}. The source code will be released in an upcoming version of the FATE framework \cite{webankfate}.

At each iteration, the communication costs of SGD are $3|\mathcal{S}|$ encrypted numbers between party A and party B, and $2n$ encrypted numbers between party A\&B and the coordinator. With our quasi-Newton framework, the communication costs become $3|\mathcal{S}|+2|\mathcal{S}_H|/L$ encrypted numbers between party A and party B, and $(2+1/L)n$ encrypted numbers between party A\&B and the coordinator. By choosing $|\mathcal{S}_H|\leq |\mathcal{S}|$, the presented quasi-Newton method introduces no more than $1/L$ additional communication costs at per communication round compared with \cite{hardy2017private}.

\section{Experiments and Conclusion}
We conduct numerical experiments on two credit scoring data sets to test the advantages of our system over the mini-batch SGD method based system in \cite{hardy2017private}. \textbf{Credit} 1 \cite{dataset:defaultcredit}:  It consists of $30000$ data instances and each instance has $n=25$ attributes; 2) \textbf{Credit} 2 \cite{dataset:givecredit}
: It contains $150000$ data instances and each with $10$ attributes. By splitting each data set into two parts vertically, each party holds a subset of features and party $B$ also holds the labels. We randomly choose $80\%$ data instances as the training set and the remaining $20\%$ as the test set. We choose $\mathcal{S}_H=\mathcal{S}$ and $L=4$ in all simulations and each algorithm stops when the loss between two consecutive epochs is less than $10^{-5}$. The number of epochs, the training loss and the area under the curve (AUC) of the receiver operating characteristics (ROC) curve on the test set are shown in table \ref{tab:credit}. Numerical results demonstrate that the proposed system requires less communication overhead than the first-order SGD based framework.

\begin{table}[h]
  \caption{Numerical Results on Two Public Data Sets}
  \label{tab:credit}
  \centering
  \begin{tabular}{cccccccc}
    \toprule
   \multirow{2}{*}{Batch Size} & \multirow{2}{*}{Method} &\multicolumn{3}{c}{\textbf{Credit 1}} & \multicolumn{3}{c}{\textbf{Credit 2}}   \\\cmidrule(r){3-5}\cmidrule(r){6-8}
   &  & Epochs     & Loss & AUC & Epochs     & Loss & AUC  \\\midrule
      \multirow{2}{*}{1000} &  SGD     & 12       & 0.496218 & 0.7224  & 12       & 0.314555 & 0.7033  \\
     & Proposed     & 3       & 0.496600 & 0.7222  & 4       & 0.314643 & 0.7061  \\\cmidrule(r){2-8}
     \multirow{2}{*}{3000} &  SGD     & 18       & 0.496194 & 0.7219 & 14       & 0.314648 & 0.6982   \\
      & Proposed     & 12       & 0.496317 & 0.7225  & 6       & 0.314490 & 0.7077  \\
     \bottomrule
  \end{tabular}
\end{table}

In this paper, we consider the communication challenges in vertical federated learning problem with two data providers for learning a logistic regression model collaboratively. We propose to use a quasi-Newton method to reduce the number of communication rounds. With the additively homomorphic encryption scheme, two data providers compute an encrypted gradient by exchanging encrypted intermediate values, and an additional vector every $L$ iterations for updating the curvature information. Numerical experiment demonstrate that our method considerably reduces the number of communication rounds with a little additional communication cost per round.

\bibliographystyle{unsrt}

\begin{thebibliography}{10}

\bibitem{yang2019federated}
Qiang Yang, Yang Liu, Tianjian Chen, and Yongxin Tong.
\newblock Federated machine learning: Concept and applications.
\newblock {\em ACM Transactions on Intelligent Systems and Technology (TIST)},
  10(2):12, 2019.

\bibitem{albrecht2016gdpr}
Jan~Philipp Albrecht.
\newblock How the {GDPR} will change the world.
\newblock {\em Eur. Data Prot. L. Rev.}, 2:287, 2016.

\bibitem{wiki:facebookdatabreach}
{Wiki}.
\newblock Data breach --- {Wikipedia}{,} the free encyclopedia.
\newblock
  \url{https://en.wikipedia.org/w/index.php?title=Data_breach&oldid=912247856},
  2019.

\bibitem{konevcny2016federated}
Jakub Kone{\v{c}}n{\`y}, H~Brendan McMahan, Felix~X Yu, Peter Richt{\'a}rik,
  Ananda~Theertha Suresh, and Dave Bacon.
\newblock Federated learning: Strategies for improving communication
  efficiency.
\newblock {\em arXiv preprint arXiv:1610.05492}, 2016.

\bibitem{mcmahan2017communication}
Brendan McMahan, Eider Moore, Daniel Ramage, Seth Hampson, and Blaise~Aguera
  y~Arcas.
\newblock Communication-efficient learning of deep networks from decentralized
  data.
\newblock In {\em Artificial Intelligence and Statistics}, pages 1273--1282,
  2017.

\bibitem{yang2018federated}
Kai Yang, Tao Jiang, Yuanming Shi, and Zhi Ding.
\newblock Federated learning via over-the-air computation.
\newblock {\em arXiv preprint arXiv:1812.11750}, 2018.

\bibitem{cheng2019secureboost}
Kewei Cheng, Tao Fan, Yilun Jin, Yang Liu, Tianjian Chen, and Qiang Yang.
\newblock Secureboost: A lossless federated learning framework.
\newblock {\em arXiv preprint arXiv:1901.08755}, 2019.

\bibitem{hardy2017private}
Stephen Hardy, Wilko Henecka, Hamish Ivey-Law, Richard Nock, Giorgio Patrini,
  Guillaume Smith, and Brian Thorne.
\newblock Private federated learning on vertically partitioned data via entity
  resolution and additively homomorphic encryption.
\newblock {\em arXiv preprint arXiv:1711.10677}, 2017.

\bibitem{nocedal2006numerical}
Jorge Nocedal and Stephen Wright.
\newblock {\em Numerical optimization}.
\newblock Springer Science \& Business Media, 2006.

\bibitem{schraudolph2007stochastic}
Nicol~N Schraudolph, Jin Yu, and Simon G{\"u}nter.
\newblock A stochastic quasi-newton method for online convex optimization.
\newblock In {\em Artificial intelligence and statistics}, pages 436--443,
  2007.

\bibitem{byrd2016stochastic}
Richard~H Byrd, Samantha~L Hansen, Jorge Nocedal, and Yoram Singer.
\newblock A stochastic quasi-newton method for large-scale optimization.
\newblock {\em SIAM Journal on Optimization}, 26(2):1008--1031, 2016.

\bibitem{moritz2016linearly}
Philipp Moritz, Robert Nishihara, and Michael Jordan.
\newblock A linearly-convergent stochastic {L-BFGS} algorithm.
\newblock In {\em Artificial Intelligence and Statistics}, pages 249--258,
  2016.

\bibitem{paillier1999public}
Pascal Paillier.
\newblock Public-key cryptosystems based on composite degree residuosity
  classes.
\newblock In {\em International Conference on the Theory and Applications of
  Cryptographic Techniques}, pages 223--238. Springer, 1999.

\bibitem{aono2016scalable}
Yoshinori Aono, Takuya Hayashi, Le~Trieu~Phong, and Lihua Wang.
\newblock Scalable and secure logistic regression via homomorphic encryption.
\newblock In {\em Proceedings of the Sixth ACM Conference on Data and
  Application Security and Privacy}, pages 142--144. ACM, 2016.

\bibitem{webankfate}
WeBank.
\newblock {FATE}: An industrial grade federated learning framework.
\newblock \url{https://fate.fedai.org}, 2018.

\bibitem{dataset:defaultcredit}
{UCI Machine Learning Repository}.
\newblock default of credit card clients data set.
\newblock
  \url{https://archive.ics.uci.edu/ml/datasets/default+of+credit+card+clients},
  2017.

\bibitem{dataset:givecredit}
{Give me some credit}.
\newblock Give me some credit.
\newblock \url{https://www.kaggle.com/c/GiveMeSomeCredit/data}, 2011.

\end{thebibliography}

\newpage
\appendix

\section*{Appendix A}
We provide details of the proposed vertical federated learning framework in Algorithm \ref{algorithm:sqn}.

\begin{algorithm}
\SetKwInOut{Input}{Input}
\SetKwInOut{Output}{Output}
 \Input{$\bs{w}_0^A,\bs{w}_0^B,M,L$}
\Output{$\bs{w}^A,\bs{w}^B$}
Set $t=0,\bs{H}=\bs{I}$\\
\For{each round $k=1,\cdots,$}{
Choose a minibatch $\mathcal{S}$ \\
\uIf{$\text{mod}(k,L)\ne 0$}{
\textbf{Party A\&B:} compute $\encrypt{\textrm{loss}},\encrypt{\bs{g}}$ as equation (3) (4)\\
\textbf{Coordinator:} $\bs{w}_{k+1}=\bs{w}_k-\tilde{\bs{g}}_k$ where $\tilde{\bs{g}}_k=\eta\bs{H}\bs{g}$
}
\Else{
$t \leftarrow t+1$\\
\textbf{Party A\&B:} Choose a minibatch $\mathcal{S}_H$ \\
compute $\encrypt{\textrm{loss}},\encrypt{\bs{g}},\encrypt{\bs{v}_t}$ as equation (3) (4) (6)\\
\textbf{Coordinator:} $\bs{w}_{k+1}=\bs{w}_k-\tilde{\bs{g}}_k$ where $\tilde{\bs{g}}_k=\eta\bs{H}\bs{g}$ \\
$\bs{s}_t=\sum_{i=k-L+1}^{k}\tilde{\bs{g}}_i/L-\sum_{i=k-2L+1}^{k-L}\tilde{\bs{g}}_i/L$\\
\If{$t>1$}{
$\bs{H}\leftarrow (\bs{s}_t^T\bs{v}_t)/(\bs{v}_t^T\bs{v}_t) \bs{I},\tilde{m}=\min\{M,t\}$ \\
\For{$j=t-\tilde{m}+1,\cdots,t$}{
$\rho_j=1/(\bs{v}_j^Ts_j)$\\
$\bs{H}\leftarrow (\bs{I}-\rho_j \bs{s}_j\bs{v}_j^T)\bs{H}(\bs{I}-\rho_j \bs{v}_j\bs{s}_j^T)+\rho_j \bs{s}_j \bs{s}_j^T$
}
}
$\tilde{w}_t=0$
}
}
\caption{A Quasi-Newton Framework for Vertical Federated Learning}
\label{algorithm:sqn} 
\end{algorithm}

\end{document}